\documentclass[runningheads]{llncs}

\usepackage[mobile]{eccv}

\usepackage{eccvabbrv}

\usepackage{verbatim}
\usepackage{graphicx}
\usepackage{booktabs}
\usepackage{multirow}
\usepackage{amsmath}
\usepackage{bbding}
\usepackage[accsupp]{axessibility}  %

\usepackage[pagebackref,breaklinks,colorlinks,citecolor=eccvblue]{hyperref}

\usepackage[capitalize]{cleveref} %
\crefname{figure}{Fig.}{Fig.}
\crefname{table}{Tab.}{Tab.}
\crefname{equation}{Eq.}{Eq.}
\crefname{section}{Sec.}{Sec.}

\usepackage{orcidlink}

\newcommand{\methodname}{\textbf{ReMamber}\xspace}

\begin{document}

\title{ReMamber: Referring Image Segmentation with Mamba Twister} 

\titlerunning{ReMamber: Referring Image Segmentation with Mamba Twister}

\author{Yuhuan Yang*\inst{1} \and
Chaofan Ma*\inst{1} \and
Jiangchao Yao\inst{1} \and
Zhun Zhong$^\text{\Envelope}$\inst{2} \and
Ya Zhang\inst{1} \and
Yanfeng Wang$^\text{\Envelope}$ \inst{1}}
\authorrunning{Y.~Yang, C.~Ma, J.~Yao, Z.~Zhong, Y.~Zhang, and Y.~Wang}

\institute{Shanghai Jiao Tong University
\and
University of Nottingham \\
\email{\{yangyuhuan,chaofanma,sunarker,ya\_zhang,wangyanfeng622\}@sjtu.edu.cn} \\
\email{zhunzhong007@gmail.com}
}

\maketitle

\begin{abstract}

Referring Image Segmentation~(RIS) leveraging transformers has achieved great success on the interpretation of complex visual-language tasks. 
However, the quadratic computation cost makes it resource-consuming in capturing long-range visual-language dependencies.
Fortunately, Mamba addresses this with efficient linear complexity in processing. 
However, directly applying Mamba to multi-modal interactions presents challenges, primarily due to inadequate channel interactions for the effective fusion of multi-modal data. 
In this paper, we propose \methodname, a novel RIS architecture that integrates the power of Mamba with a multi-modal \textit{Mamba Twister} block. 
The Mamba Twister explicitly models image-text interaction, and fuses textual and visual features through its unique channel and spatial \textit{twisting mechanism}.
We achieve competitive results on three challenging benchmarks with a simple and efficient architecture. Moreover, we conduct thorough analyses of \methodname and discuss other fusion designs using Mamba.
These provide valuable perspectives for future research.
The code has been released at: \url{https://github.com/yyh-rain-song/ReMamber}.

\keywords{Referring Image Segmentation (RIS) \and Multi-Modal Understanding \and Mamba Architecture}
\end{abstract}

\begin{figure}[t]
    \centering
    \includegraphics[width=\linewidth]{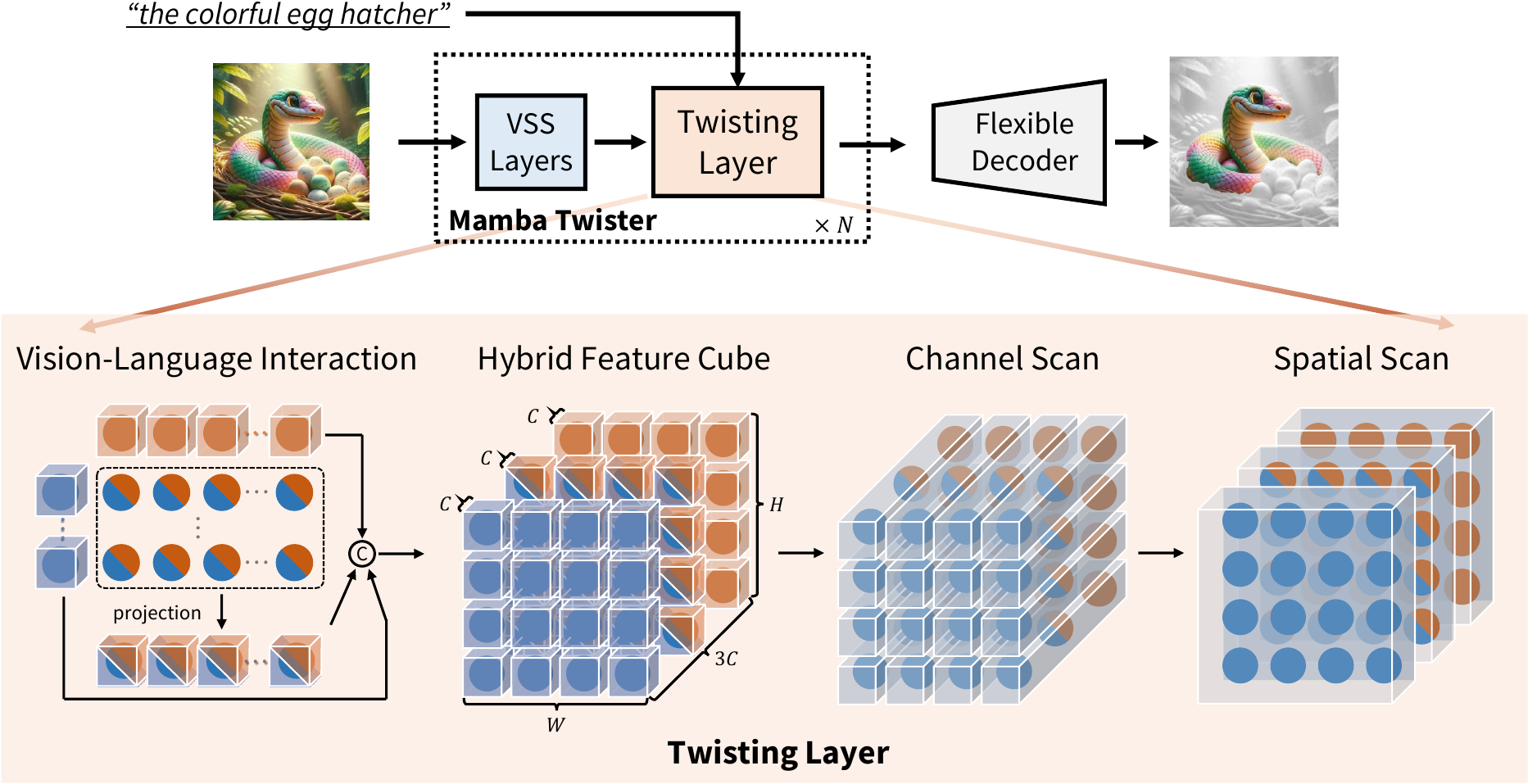}
    \caption{We propose \methodname, a novel \underline{re}ferring segmentation architecture with \underline{Mamb}a twist\underline{er}. It consists of several \textit{Mamba Twister block}. Each block contains several \textit{visual state space (VSS) layers} and a \textit{Twisting layer}. The Twisting layer first calculates the interaction between image and text, and then forms a hybrid feature cube. Finally, it ``twists'' the feature cube using the Channel and Spatial Scan along each dimension.}
    \label{fig:teaser}
    \vspace{-1.5em}
\end{figure}

\section{Introduction}
\label{sec:intro}

Referring Image Segmentation (RIS) is a crucial yet challenging task in the area of multi-modal understanding~\cite{li2018referring, liu2017recurrent, hu2016segmentation}. Unlike ordinary image segmentation, RIS involves the identification and segmentation of specific objects in image based on the textual descriptions. This thereby requires the model to be capable of understanding vision-language interactions, which is the core challenge of RIS.

Thanks to the powerful attention mechanisms, it has achieved great success for RIS by leveraging transformers to promote the exact recognition of multi-modal information.
For example, existing works have designed transformer decoder~\cite{kamath2021mdetr,ding2021vision} or transformer encoder-decoder~\cite{yang2022lavt, kim2022restr, wang2022cris, ma2022fusioner,yang2023multimodal, ma2023attrseg} to comprehensively fuse visual and linguistic features, which have achieved great progress.
Nonetheless, it is a quadratic increase in both computation and memory when applying full attention in transformers~\cite{vaswani2017attention, dosovitskiy2020image}.
{This leads to the limitation for resource-intensive scenarios, for example, in capturing long-range visual-language dependencies. And this is particularly important in the context of large-size images with long textual descriptions.}

Fortunately, recent advances of State Space Models (SSMs)~\cite{gu2021efficiently, smith2022simplified, fu2022hungry, gu2023mamba} has emerged as promising architectures for solving the above issue.
Specifically, Mamba~\cite{gu2023mamba} marks a significant advancement for efficient training and inference with linear complexity, which has been incorporated into various visual tasks. However, current efforts primarily pay attention to \textit{single-modality} settings, such as image classification~\cite{zhu2024vision, liu2024vmamba}, biomedical image segmentation~\cite{ma2024u,xing2024segmamba,ruan2024vm,liu2024swin}, low-level vision~\cite{zheng2024u,he2024pan}, and point cloud analysis~\cite{liang2024pointmamba,zhang2024point}.
In this paper, we pioneer the exploration of Mamba in \textit{multi-modal} RIS setting, 
and identify that the prevalent multi-modality token splicing method in transformers~\cite{kim2022restr,ding2021vision,yang2022lavt} is no longer effective.  
Since in Mamba, a fundamental deficiency exists: the interactions are insufficient between channels of different tokens~\cite{gu2023mamba}, which adversely affects the fusion of multi-modal information in RIS. 

To address this dilemma, we propose \methodname, a novel \underline{re}ferring segmentation architecture with \underline{Mamb}a twist\underline{er}. 
As shown in \cref{fig:teaser}, \methodname comprises several \textit{Mamba Twister blocks}, which allow the model to be ``aware'' of the textual context at every spatial location.
Each block consists of several \textit{visual state space (VSS) layers} and a \textit{Twisting layer}.
Specifically, the VSS layers initially {extract visual features}, and the Twisting layer injects the textual information into the visual modality.
The \textit{Twisting layer} is structured into three critical components.
\textbf{(1)} A \textit{vision-language interaction operation} is designed to explicitly capture the fine-grained interactions between modalities.
This is achieved by evaluating the similarity between visual and textual tokens then mapping them into a shared feature space, thus generating a multi-modal feature.
\textbf{(2)} A \textit{hybrid feature cube} is then created by concatenating the visual feature, multi-modal feature, and global textual feature. 
This ensures that each visual token receives uniform influence from both local and global contexts, preventing the overshadowing of subtle textual cues by predominant visual features.
\textbf{(3)} Lastly, to address the inadequate interaction within channels in Mamba, a \textit{twisting mechanism} is deployed. 
This process ``twists'' the feature cube channel- and spatial-wise, thereby enhancing interaction within and across modalities. It is accomplished through two consecutive SSMs scanning along channel and spatial dimensions, respectively.

To sum up, the key contributions of our research are highlighted as follows:

\noindent\textbf{$\bullet$} We pioneer the exploration of Mamba in referring image segmentation (RIS), demonstrating Mamba's significant potential for multi-modal understanding.

\noindent\textbf{$\bullet$} We design a novel framework, \methodname, that mainly contains several Mamba Twister blocks. 
This design effectively captures vision-language interactions using the ``twisting mechanism''.

\noindent\textbf{$\bullet$} We achieve competitive results on three challenging benchmarks. Moreover, we conduct thorough analyses of \methodname and discuss other vision-language fusion designs using Mamba.
These provide valuable perspectives for future research.

\section{Related Work}

\subsection{Referring Image Segmentation}

Referring image segmentation (RIS) aims to segment the entities in the image following natural language instruction.
Early approaches~\cite{li2018referring, liu2017recurrent, hu2016segmentation, margffoy2018dynamic} leveraged RNNs or LSTMs to encode linguistic representations, 
and CNNs to extract spatial features from the image at varying levels.
These disparate modalities were then integrated using the multi-modal LSTM~\cite{li2018referring, liu2017recurrent, hu2016segmentation, margffoy2018dynamic, chen2019see, jing2021locate}, attention mechanism~\cite{ye2019cross, shi2018key, hu2020bi, feng2021encoder}, cycle-consistency~\cite{chen2019referring}, and graph convolution~\cite{huang2020referring}.

Recent trends have shifted towards leveraging transformers for enhanced capturing and fusion of vision-language modalities.
MDETR~\cite{kamath2021mdetr} and VLT~\cite{ding2021vision}
design a transformer decoder for fusing linguistic and visual features.
LAVT~\cite{yang2022lavt} adopts Swin Transformer as the visual backbone and incorporate vision-language fusion modules within the visual encoder's final layers.
Similar strategies are employed by ReSTR~\cite{kim2022restr} and CRIS~\cite{wang2022cris}, which utilize dual transformer encoders for initial modality encoding, followed by feature fusion through a multi-modal transformer encoder or decoder.
Other models like PolyFormer~\cite{liu2023polyformer}, SeqTr~\cite{zhu2022seqtr} and \cite{qu2023learning} also adopt a multi-modal transformer for vision-language fusion, but they output masks as sequences of contour points.
Meanwhile, GRES~\cite{liu2023gres} and CGFormer~\cite{tang2023contrastive} consider transformer queries as region proposals, and regarding segmentations as proposal-level classification problems.
Distinct from all existing works, we introduce a pioneering multi-modal architecture named \methodname, based on Mamba~\cite{gu2023mamba}. 
This novel approach underscores the untapped potential of Mamba in advancing the field of referring image segmentation.

\subsection{State Space Models and Visual Applications}

State space models (SSMs), originally derived from control theory, have been effectively combined with deep learning to model the long-range dependencies.
Early works like LSSL~\cite{gu2021combining} show potential when combined with HiPPO~\cite{gu2020hippo} initialization.
S4~\cite{gu2021efficiently} was designed to diminish both computational and memory demands associated with state representations.
It scales linearly with sequence length, offering a notable advantage over CNNs and transformers.
Building on S4, S5~\cite{smith2022simplified} introduces the efficient parallel scan and MIMO SSM, further refining the approach.
Later, H3~\cite{fu2022hungry} expanded on these foundations, achieving competitive performance with transformers in language modeling.
Recently, Mamba~\cite{gu2023mamba} marked a significant advancement with its linear-time inference and efficient training, incorporating a selection mechanism and hardware-aware algorithms building upon prior works~\cite{mehta2022long, gupta2022diagonal, gu2022parameterization}.

With the demonstrated success of SSMs like Mamba in language modeling, researchers have begun to investigate their applicability to visual tasks. 
Works such as ViS4mer~\cite{islam2022long}, Selective S4~\cite{wang2023selective} and TranS4mer~\cite{islam2023efficient} directly use the S4's ability to capture long-range sequences for understanding inter-frame relations in video classification and detection.
However, they still employ Vision Transformers for intra-frame feature extraction. 
More recently, Vim~\cite{zhu2024vision} and VMamba~\cite{liu2024vmamba} have shown promising results as fully Mamba-based vision backbones for image classification, detection, and segmentation.
At the same time, several studies have explored Mamba-based architecture on various vision tasks, such as biomedical image segmentation~\cite{ma2024u,xing2024segmamba,ruan2024vm,liu2024swin}, low-level vision~\cite{zheng2024u,he2024pan}, and point cloud analysis~\cite{liang2024pointmamba,zhang2024point}.
Nevertheless, all above efforts primarily focus on \textit{single-modality} vision tasks. 
In this paper, 
we pioneer the exploration of Mamba-base architecture in \textit{multi-modality} RIS task.

\section{Methods}

This paper aims to develop a multi-modal Mamba-based architecture for RIS task. 
In~\cref{sec:preliminary}, we give the preliminary.
From~\cref{sec:overview} to~\cref{sec:Mamba_twister}, we describe our architecture from coarse to fine.
In~\cref{sec:CI_variants}, we introduce other three variants besides our design.
And in~\cref{sec:training} we detail the training process.

\subsection{Preliminary: State Space Model}
\label{sec:preliminary}
State Space Model (SSM) is a sequence model inspired by continuous systems. 
It is designed to capture a mapping relationship between two functions or sequences, expressed as $x(t) \in \mathbb{R} \mapsto y(t) \in \mathbb{R}$, through a hidden state $h(t) \in \mathbb{R}^N$.
The evolution of the hidden state over time is governed by specific parameters $\mathbf{A} \in \mathbb{R}^{N \times N}$, which directs the state evolution, and $\mathbf{B} \in \mathbb{R}^{N}, \mathbf{C} \in \mathbb{R}^{N}$, which performs the projection. Formally, SSM can be described by the following equations:
\begin{equation}
\begin{aligned}
\label{eq:continuous_system}
h'(t) &= \mathbf{A} h(t)+\mathbf{B} x(t), \\ 
y(t) &= \mathbf{C} h(t) .
\end{aligned}
\end{equation}

Discretizing this system introduces a timescale parameter $\boldsymbol{\Delta} t$, which transforms the continuous parameters $\mathbf{A}, \mathbf{B}$ into their discrete counterparts $\overline{\mathbf{A}}, \overline{\mathbf{B}}$. 
A common way for this transformation is the zero-order hold (ZOH) approach, defined as:
\begin{equation}
\begin{aligned}
\overline{\mathbf{A}}&=\exp (\boldsymbol{\Delta} \mathbf{A}), \\
\overline{\mathbf{B}}&=(\boldsymbol{\Delta} \mathbf{A})^{-1}(\exp (\boldsymbol{\Delta} \mathbf{A})-\mathbf{I}) \cdot \boldsymbol{\Delta} \mathbf{B}.
\end{aligned}
\end{equation}
Then, the discretized version of this system can be represented as:
\begin{equation}
\begin{aligned}
h_t &=\overline{\mathbf{A}} h_{t-1}+\overline{\mathbf{B}} x_t, \\
y_t &=\mathbf{C} h_t .
\end{aligned}
\end{equation}
Mamba~\cite{gu2023mamba}, as a variant of SSM, releases the linearity constraint of the original SSM described in~\cref{eq:continuous_system}, by making $\mathbf{B}$ and $\mathbf{C}$ to be input-dependent, whereas still maintaining the linear complexity during forward process.

\subsection{Architecture Overview}
\label{sec:overview}
\begin{figure}[t]
    \centering
    \includegraphics[width=\linewidth]{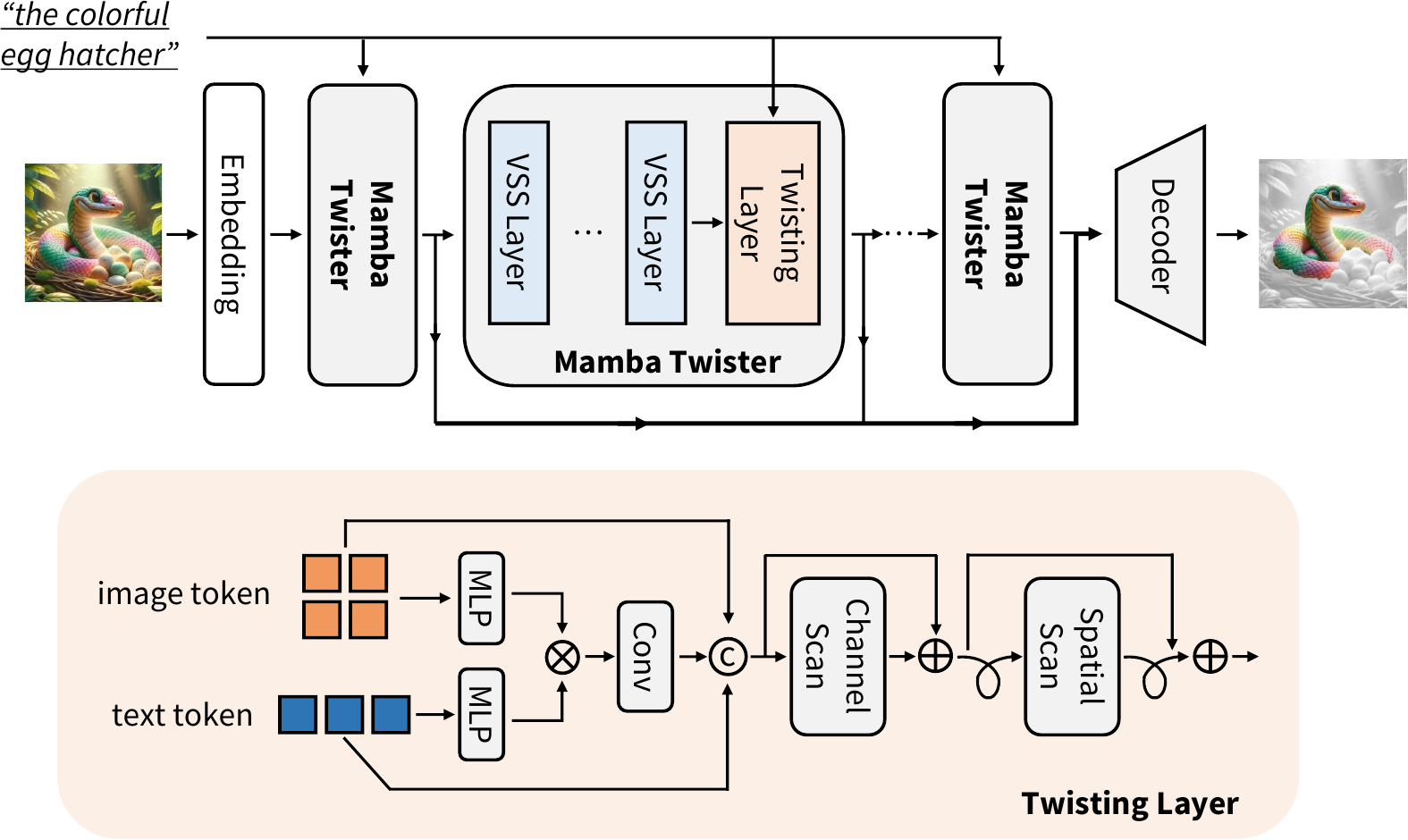}
    \caption{\textbf{Overview architecture of our \methodname.} 
    The basic block for \methodname is the \textit{Mamba Twister block}. It consists of several visual state space \textit{(VSS) layers} and a \textit{Twisting layer}. The Twisting layer first constructs hybrid feature cube from text, image, and multi-modal features via channel concatenation. Then, it ``twists'' the cube by Channel Scan and Spatial Scan. We extract intermediate features after each Mamba Twister block, and feed it into a flexible decoder for final segmentation.
    }
    \label{fig:architecture}
    \vspace{-1.5em}
\end{figure}

The pivotal aspect of Referring Image Segmentation (RIS) is the correspondence between image and textual input.
To achieve this, we propose \methodname.
\cref{fig:architecture} shows the overview of our architecture. 
The basic block for \methodname is the \textit{Mamba Twister} block.
It is a multi-modal fusion block that takes both visual and textual feature as input, and outputs the fused multi-modal feature representation.
We extract the intermediate feature after each Mamba Twister block, and then feed it into a flexible decoder to generate the final segmentation mask.
The decoder is task-invariant, so any downstream architecture can be applied.
Below in \cref{sec:Mamba_Twister_Block}, we will first introduce the key component: Mamba Twister block.

\subsection{Mamba Twister Block}
\label{sec:Mamba_Twister_Block}
As shown in~\cref{fig:architecture}, the Mamba Twister Block consists of several \textit{visual state space (VSS) layers} and a \textit{Twisting layer}.
The \textit{VSS layer} is designed to process features in the spatial domain. 
It treats the input feature as a series of image tokens and aims to discern the spatial relationships among these tokens.
The \textit{\textit{Twisting layer}} aims to inject text condition into the image feature, thus guiding the transformation of image feature.
It begins by condensing the textual sequence to compute the cross-correlation between the image and textual modalities.
Subsequently, it disseminates this information across each image feature patch via a twisting operation.
We'll detail the introduction of VSS layer in~\cref{sec:VSSM} and the Twisting layer in~\cref{sec:Mamba_twister}.

\subsection{VSS Layer for Spatial Data Processing}
\label{sec:VSSM}

Since the SSM is initially designed for processing temporal or causal data, which can not effectively process non-causal data types, \ie, 2-dimensional image in our case. 
To solve this issue, we adopt the Cross-Scan-Module proposed in~\cite{liu2024vmamba}, which unfolds image patches into sequences and then scans them in four distinct directions, ensuring that the information from all pixels is integrated during feature transformation.
We replace the scanning operation in vanilla Mamba block with CSM, forming a visual state space (VSS) layer for our spatial feature transformation.

\subsection{Twisting Layer for Multi-modal Fusion}
\label{sec:Mamba_twister}

Here we introduce the detailed design of our \textit{Twisting layer}. 
Intuitively, the Twisting layer first constructs a feature cube by arranging information from different modalities orderly. 
Then, pieces of information intertwine with each other when the cube is \textit{twisted}
through two SSM layers along different axis, thereby achieving modality fusion.
Formally, given image feature $\mathbf{F}_i\in\mathbb{R}^{h\times w\times C_i}$ and text feature $\mathbf{F}_t\in\mathbb{R}^{L\times C_t}$, the Mamba twister aims to learn a function $(\mathbf{F}_{i}, \mathbf{F}_t) \mapsto \tilde{\mathbf{F}}\in \mathbb{R}^{h\times w\times C_o}$ by first forming a hybrid multi-modal feature cube and then conduct two SSM scans on it. $C_i, C_t$ and $C_o$ are the dimensions of visual, textual, and output feature; $h, w$ and $L$ are height, weight, and length.

\subsubsection{Forming the Hybrid Feature Cube. }
To explicitly build the image-text correspondence, we design a novel \textit{vision-language interaction operation} and then form a \textit{hybrid feature cube}.
This allows more effective modality fusion.

To be specific, we formulated the vision-language interaction operation into two manners: \textbf{global interaction} and \textbf{local interaction}.
The \textbf{global interaction} treats the text sequence as a whole, which means that all image patches should be aware of the semantic meaning of the text expression.
We pool a global representation $\mathbf{F}_t^\mathrm{CLS}\in\mathbb{R}^{C_t}$ from the text sequence $\mathbf{F}_t$ to convey this information.
The $\mathbf{F}_t^\mathrm{CLS}$ is further expanded to the same size as the image feature.
Formally:
\begin{equation}
    \tilde{\mathbf{F}}_t = \mathrm{Expand}(\mathbf{F}_t^\mathrm{CLS})\in\mathbb{R}^{h\times w\times C_t},
\end{equation}
where $\mathrm{Expand}(\cdot)$ denotes the operation that expands the input tensor to the same size as the image feature.

However, a global representation may not be enough to capture the intricate relationships between different modalities.
For example, some words such as color, shape or location may be more relevant to certain image patches than others.
The \textbf{local interaction} aims to capture such correlation in a fine-grained level. Formally, we calculate the local interaction map $\mathbf{F}_c$ using matrix multiplication:
\begin{equation}
    \mathbf{F}_c = \mathbf{F}_i\mathbf{W}_i \cdot (\mathbf{F}_t\mathbf{W}_t)^T\in\mathbb{R}^{h\times w\times L},
    \label{eq:fc}
\end{equation}
where $\mathbf{W}_i\in\mathbb{R}^{C_i\times C_c}$ and $\mathbf{W}_t\in\mathbb{R}^{C_t\times C_c}$ are learnable parameters.
To enhance feature processing in high dimension, we use a single convolutional layer to transform $\mathbf{F}_c$ into $\tilde{\mathbf{F}}_c \in \mathbb{R}^{h\times w\times C_c}$.
We then concatenate $\mathbf{F}_i$, $\tilde{\mathbf{F}}_t$, and $\tilde{\mathbf{F}}_c$ along the channel dimension to form the hybrid feature cube.
Formally:
\begin{equation}
    \mathbf{F}_\mathrm{cube} = [\mathbf{F}_i,\tilde{\mathbf{F}}_t,\tilde{\mathbf{F}}_c]\in\mathbb{R}^{h\times w\times (C_i+C_t+C_c)}.
    \label{eq:hybrid_cube}
\end{equation}

\subsubsection{Twisting the Hybrid Feature Cube. }
Generally, the scanning operation for vanilla SSM is almost independent across channels~\cite{gu2023mamba}.
This is not enough for feature communication, especially when information from different modalities is arranged along the channel dimension, \ie, our hybrid feature cube.

To foster the interactions \textbf{within} and \textbf{across} modalities, here we design a \textbf{twisting mechanism} to ``twist'' the hybrid feature cube along different axis.
Specifically, it is composed of the Channel Scan and Spacial Scan successively.
The \textbf{Channel Scan} first treats the channel as an ordered sequence, and learns to communicate cross channels, thus foster the modality fusion.
Then, the \textbf{Spatial Scan} operates on the spatial dimension, and learns to communicate information cross patches within each modality separately.
Formally, this process can be written as:
\begin{equation}
    \mathbf{F}_\text{out} %
    = \mathrm{SSM}_{\mathrm{spatial}}\left(\mathrm{SSM}_{\mathrm{channel}}\left(\mathbf{F}_\mathrm{cube}\right)\right),
\end{equation}
where $\mathrm{SSM}_{\mathrm{channel}}$ refers to the Channel Scan. It is an SSM layer that treats the concatenated feature as an ordered sequence through channel dimension, and presents 1-D selective scan along the channel. While $\mathrm{SSM}_{\mathrm{spatial}}$ denotes the Spatial Scan. 
It is an VSS layer and presents 2D selective scan along the two spatial dimension.
$\mathbf{F}_\text{out}$ is then feed into the next layer for further process.

\subsection{Discussion: Other Variants of Modality Fusion Designs}
\label{sec:CI_variants}
\begin{figure}[t]
    \centering
    \includegraphics[width=0.95\linewidth]{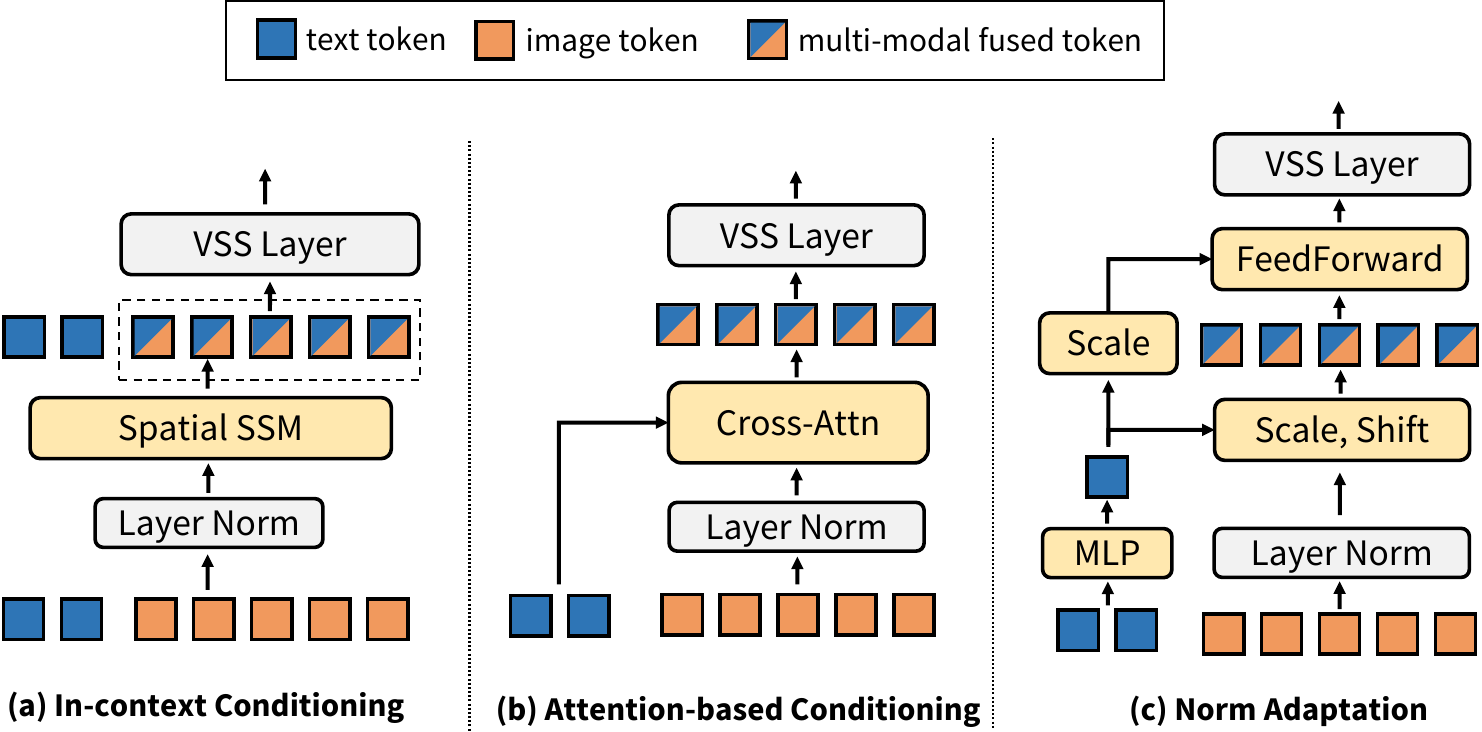}
    \caption{\textbf{Other multi-modal fusion designs.}  (a) \textbf{In-context Conditioning} appends text tokens ahead of image tokens.
     (b) \textbf{Attention-based Conditioning} utilizes cross-attention mechanism for modality fusion.
     (c) \textbf{Norm Adaptation} learns a scale and bias for the model's normalization layers.}
    \label{fig:CI_variants}
    \vspace{-1.5em}
\end{figure}

Besides the proposed Twisting layer, we also explore other variants of multi-modal fusion designs. 
\cref{fig:CI_variants} shows the structure of three other different variants.
We here discuss these variants, and later in \cref{sec:CI_ablation}, we analyze the results for these variants.

\noindent\textbf{In-context Conditioning.}
As shown in \cref{fig:CI_variants} (a), in this variant, we simply append the text sequence before the image feature, forming a longer sequence than originally being processed. 
In this way, the model is able to fuse image features with the previous text as a context.
This is the most straightforward way to allow the model aware of the text condition.

\noindent\textbf{Attention-based Conditioning. }
Attention mechanism is usually a strong baseline for sequence modeling.
Here we utilize the cross-attention mechanism for two modality fusion, as shown in \cref{fig:CI_variants} (b),
To be specific, we use the image feature as query, and the text feature as key and value, hoping to capture the correlation between image and text.

\noindent\textbf{Norm Adaptation. }
In this variant, we integrate the text input by adapting the scale and bias after norm layer using FiLM~\cite{perez2018film}.
We first pool a global representation from text sequence, and then use a linear projection to transform it into the scale and shift in layer normalization.

\subsection{Model Training}
\label{sec:training}
The proposed \methodname is a general framework for multi-modal fusion. Therefore, there is no restriction on the downstream decoder design nor the loss function. Here we use a simple convolution-based decoder for simplicity.
The entire network is trained in an end-to-end manner.

\section{Experiments}
\subsection{Datasets and Metrics}

\textbf{Dataset.} To assess the efficacy of our proposed approach, we executed experiments across three widely recognized datasets tailored for the referring image segmentation (RIS) task: RefCOCO, RefCOCO+~\cite{referitgame}, and G-Ref~\cite{referitgame,gref,2016VarunModeling}.
{RefCOCO}~\cite{referitgame} is a prominent dataset in the RIS domain, comprising 19,994 images and 142,210 referring expressions associated with 50,000 objects derived from the MSCOCO~\cite{mscoco} dataset through a two-player game. 
{RefCOCO+} enhances the challenge by excluding expressions with absolute location references, featuring 141,564 expressions for 49,856 objects across 19,992 images. 
{G-Ref}~\cite{gref,2016VarunModeling} enriches the dataset diversity with 104,560 referring expressions for 54,882 objects in 26,711 images, showcasing an average sentence length of 8.4 words with a heightened focus on location and appearance descriptions.

\noindent\textbf{Evaluation Metrics.} In line with preceding studies~\cite{cgformer}, we adopt mIoU and oIoU as the main metrics. 
Furthermore, we incorporate the metric Precision@X (X$\in\{50,60,70,90\}$) for a more comprehensive evaluation, whereas Percision@X means the percentage of test images with an IoU score high than X$\%$.

\subsection{Implementation Details}

Our model architecture is developed upon the foundations of Mamba~\cite{gu2023mamba} and VMamba~\cite{liu2024vmamba}.
We adopt ImageNet pre-trained weights as initialization, and train the model in an end-to-end manner.
We set the input image resolution to $480$ when compared with the state-of-the-art methods in~\cref{tab:main_table}, following~\cite{yang2022lavt,cgformer} for fair comparison. 
For other experiments (\cref{tab:ablation_CI,tab:ablation_scan,tab:ablation_feature,tab:ablation_feature}), we set the image resolution to $256$ for faster training and ablation study without changing its property.
For details about the training settings, please refer to our code.\footnote[1]{https://github.com/yyh-rain-song/ReMamber}

\subsection{Evaluation on Other Variants of Modality Fusion Designs}
\label{sec:CI_ablation}
As discussed in~\cref{sec:CI_variants}, besides our Mamba Twister, we also provide three other variant architectures for modality fusion,
including \textbf{In-context Conditioning}, \textbf{Attention-based Conditioning} and \textbf{Norm Adaptation}.
To make a fair comparison, for all architectures, we initialize their common part (VMamba-based parameters) with the same checkpoint.
The difference is that the Twisting layer is replaced by the corresponding fusion methods. 
The parameters of fusion modules are nearly equal, \textit{i.e.}, \{103, 113, 118, 116\} KB for \{In-Context, Attention, Norm Adaptation, Twister\}.
Here we provide a comprehensive analysis between these architectures, showing the advantages and disadvantages of each variant.

\begin{table}[!t]
    \centering
    \caption{\textbf{Comparison with other modality fusion variants. }
    ``Attention'' means for ``Attention-based Conditioning'', ``In-Context'' for ``In-Context Conditioning'',  and ``Adaptation'' for ``Norm Adaptation''.
    Mamba Twister steadily outperforms other variants across all metrics and datasets, indicating its superior capability in capturing and integrating contextual information for more accurate segmentation.}
    \label{tab:ablation_CI}
    \setlength\tabcolsep{3pt}
    \resizebox{1.\linewidth}{!}{%
    \begin{tabular}{c|cccc|cccc|cccc}
    \toprule
    \multicolumn{13}{c}{\multirow{2}{*}{Dataset: RefCOCO}}\\
    \multicolumn{13}{c}{}\\
    \midrule
     &\multicolumn{4}{c|}{val}
     &\multicolumn{4}{c|}{testA}
     &\multicolumn{4}{c}{testB}\\
    Variants 
    & mIoU & oIoU & Pr@50 & Pr@70 %
    & mIoU & oIoU & Pr@50 & Pr@70 %
    & mIoU & oIoU & Pr@50 & Pr@70 %
    \\
    \midrule
    Attention &
    65.3 & 62.3 & 75.1 & 60.6 &
    67.5 & 64.4 & 79.1 & 65.4 &
    61.7 & 58.0 & 69.6 & 53.4 \\
    In-Context &
    69.1 & 65.9 & 79.7 & 67.5 &
    71.2 & 68.9 & 82.2 & 71.8 &
    66.2 & 62.8 & 75.0 & 61.7 \\
    Adaptation &
    70.2 & 67.0 & 80.7 & 69.6 &
    72.3 & 70.2 & 83.1 & 73.2 &
    66.8 & 62.8 & 75.1 & 63.1 \\
    \textbf{Mamba Twister} &
    \bf 71.6 & \bf 68.4 & \bf 82.1 & \bf 70.9 &
    \bf 73.3 & \bf 71.5 & \bf 84.8 & \bf 74.5 &
    \bf 68.4 & \bf 64.5 & \bf 77.1 & \bf 64.9 \\
    \bottomrule
    \multicolumn{13}{c}{\multirow{2}{*}{Dataset: RefCOCO+}}\\
    \multicolumn{13}{c}{}\\
    \midrule
     &\multicolumn{4}{c|}{val}
     &\multicolumn{4}{c|}{testA}
     &\multicolumn{4}{c}{testB}\\
    Variants 
    & mIoU & oIoU & Pr@50 & Pr@70 %
    & mIoU & oIoU & Pr@50 & Pr@70 %
    & mIoU & oIoU & Pr@50 & Pr@70 %
    \\
    \midrule
    Attention &
    54.0 & 49.7 & 60.7 & 45.5 &
    58.7 & 55.8 & 68.5 & 52.9 &
    46.9 & 41.9 & 50.6 & 36.6 
    \\
    In-Context &
    58.4 & 54.0 & 66.3 & 53.9 & 
    63.2 & 59.4 & 72.2 & 60.5 & 
    51.5 & 47.5 & 56.4 & 44.5
    \\
    Adaptation &
    60.3 & 55.2 & 68.2 & 57.3 & 
    64.2 & 59.9 & 73.4 & 62.9 & 
    53.9 & 48.7 & 59.0 & 47.2
    \\
    \textbf{Mamba Twister} &
    \bf 61.6 & \bf 57.3 & \bf 70.0 & \bf 58.5 & 
    \bf 65.8 & \bf 62.1 & \bf 75.5 & \bf 64.9 & 
    \bf 54.0 & \bf 49.9 & \bf 59.4 & \bf 47.8     
    \\
    \bottomrule
    \multicolumn{13}{c}{\multirow{2}{*}{Dataset: G-Ref}}\\
    \multicolumn{13}{c}{}\\
    \midrule
     &\multicolumn{6}{c|}{val}
     &\multicolumn{6}{c}{test}\\
    Variants 
    & mIoU & oIoU & Pr@50 & \multicolumn{1}{c}{Pr@60} & Pr@70 & \multicolumn{1}{c|}{Pr@90}
    & mIoU & \multicolumn{1}{c}{oIoU} & Pr@50 & {Pr@60} & Pr@70 & \multicolumn{1}{c}{Pr@90}
    \\
    \midrule
    Attention &
    50.5 & 49.2 & 54.2 & \multicolumn{1}{c}{46.0} & 36.7 & \multicolumn{1}{c|}{8.1} &
    50.8 & \multicolumn{1}{c}{50.8} & 57.3 & 49.8 & 38.6 & 6.4
    \\
    In-Context &
    54.8 & 53.4 & 59.5 & \multicolumn{1}{c}{52.3} & 43.4 & \multicolumn{1}{c|}{13.3} &
    55.5 & \multicolumn{1}{c}{54.8} & 60.7 & 53.6 & 45.3 & 13.3 
    \\
    Adaptation &
    59.3 & 56.7 & 66.7 & \multicolumn{1}{c}{61.0} & 53.7 & \multicolumn{1}{c|}{18.4} &
    58.6 & \multicolumn{1}{c}{56.6} & 65.5 & 59.8 & 52.8 & 18.2 
    \\
    \textbf{Mamba Twister} &
    \bf 61.1 & \bf 58.0 & \bf 68.4 & \multicolumn{1}{c}{\bf62.8} & \bf 55.2 & \multicolumn{1}{c|}{\bf18.5} &
    \bf 61.2 & \multicolumn{1}{c}{\bf59.0} & \bf 69.0 & \bf 63.4 & \bf 55.5 & \bf 18.7     
    \\
    \bottomrule
    \end{tabular}}
    \vspace{-1.5em}
\end{table}

\cref{tab:ablation_CI} shows the comparison result. 
The Mamba Twister consistently outperforms the other variants across all metrics and datasets, indicating its superior capability in capturing and integrating contextual information for more accurate segmentation.

Surprisingly, despite widely used in transformers, the \textbf{Attention-based Conditioning} performs poorly in our task. 
This suggests that the cross-attention mechanism is not inherently suitable for Mamba architecture. 
This may be due to \textit{a fundamental discrepancy} between the two systems.
To be specific, 
\textbf{(1)} the Mamba model is predicted on ordered sequences that exhibit \textit{strict sequential dependencies}, where the state at a given time point $t+1$ is determined by the preceding time point $t$ and a hidden state; 
\textbf{(2)} In contrast, cross-attention mechanisms treats all tokens within a sequence \textit{equally}. 
This discrepancy may undermine the Mamba model's ability to structurally model sequences, as the cross-attention mechanism does not preserve the sequential integrity and the hierarchical dependencies essential for the model's operation.

Serving as a straightforward baseline, \textbf{In-context Conditioning} performs sub-optimal. This may because
it models image-text interaction in an implicit way. 
In our situation, the length of the image feature is much larger than the textual side, textual information may be diluted during the forward process.
Besides, this operation does not differentiate the textual information between image patches, and would be insufficient for multi-modal fusion.

Finally, \textbf{Norm Adaptation} with explicit image-text interaction modeling appears to be a strong baseline.
It achieves better performance than other two variants.
However, when calculating scale and bias, it uses only a pooled vector as textual representation instead of the entire sequence.
This potentially results in the loss of information, making it less effective compared to our Mamba Twister.

\subsubsection{Case-study Using Attention Maps. }
\begin{figure}[ht]
    \centering
    \includegraphics[width=1.\linewidth]{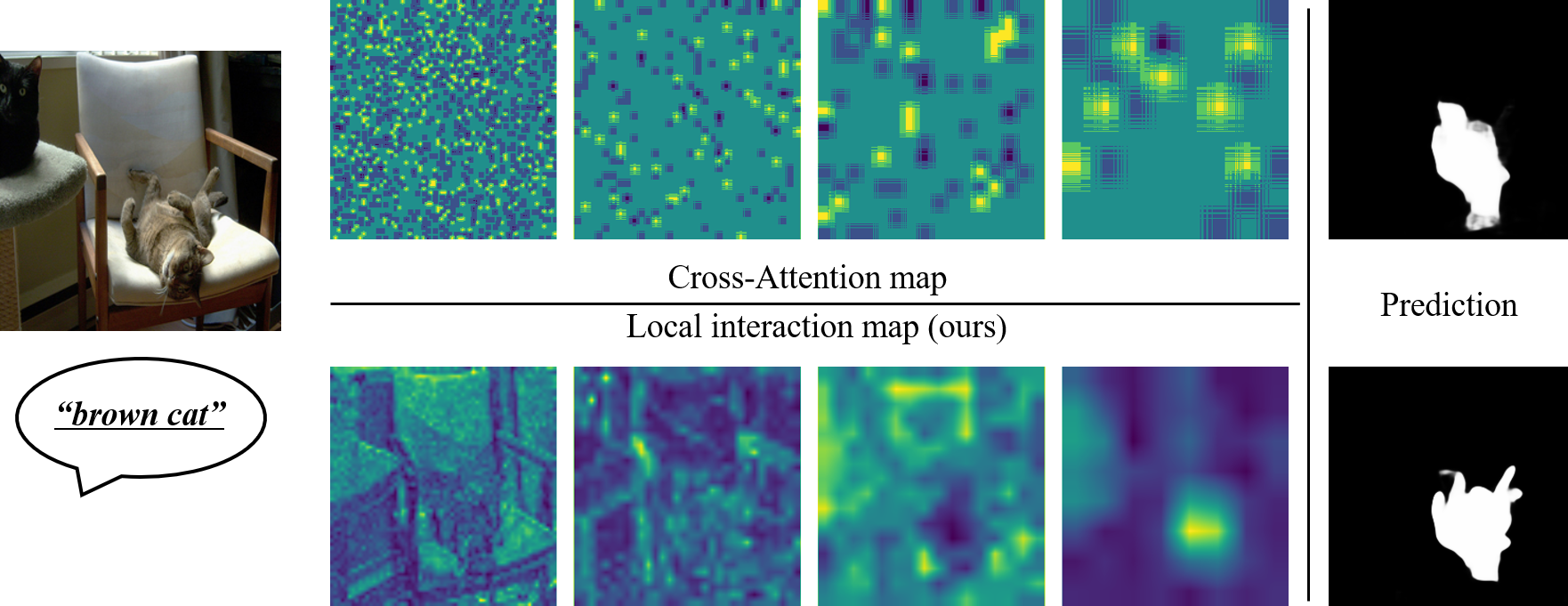}
    \caption{\textbf{Cross-Attention map (up) and our local interaction map (down) comparison. }
    Though both methods are able to predict target correctly, the cross-attention maps don't show correct image-text correlation, while ours are able to capture this relationship accurately, indicating that Mamba Twister is able to gradually fusing the two modality.}
    \label{fig:vis_cost}
    \vspace{-1.5em}
\end{figure}

In~\cref{fig:vis_cost}, we further visualized the attention maps in {Attention-based Conditioning} as well as the local interaction map in Mamba Twister defined in~\cref{eq:fc}. 
The four images in~\cref{fig:vis_cost}(b) are arranged from left to right with the network going from shallow to deep.
It can be observed that in the shallower layers, Mamba Twister focuses more on the lower-level features of the images such as edges.
As the layer goes deeper, the cost map mostly concentrating on the target object, indicating that Mamba Twister can progressively guide the image feature towards the target described by the language.
In contrast, though the Attention-based Conditioning variant is able to locate the target, its attention map performs poorly.
This may suggest that the cross-attention mechanism struggles to accurately capture the correct context information, and the two modalities are not truly fused together.

\subsection{Comparison with the State-of-the-Arts}
\begin{table}[!t]
    \centering
    \caption{\textbf{Comparison with state-of-the-art methods on RefCOCO, RefCOCO+, and G-Ref.} The metric of oIoU is reported. Best results are in bold.} %
    \label{tab:main_table}
    \setlength\tabcolsep{5pt}
    \resizebox{1.\linewidth}{!}{%
        \begin{tabular}{cc|ccc|ccc|cc}
            \toprule
            && \multicolumn{3}{c|}{RefCOCO} & \multicolumn{3}{c|}{RefCOCO+} & \multicolumn{2}{c}{G-Ref}                                                                             \\
            Model   & Backbone            & val   & testA & testB & val   & testA & testB & valU  & testU \\
            \midrule
            PCAN~\cite{chen2022position}    & ResNet-50           & 69.51 & 71.64 & 64.18 & 58.25 & 63.68 & 48.89 & 59.98 & 60.8  \\
            MAttNet~\cite{Yu_2018_CVPR} & MaskRCNN ResNet-101 & 56.51 & 62.37 & 51.7  & 46.67 & 52.39 & 40.08 & 47.64 & 48.61 \\
            RMI~\cite{liu2017recurrent}     & DeepLab ResNet-101  & 45.18 & 45.69 & 45.57 & 29.86 & 30.48 & 29.50 & -     & -     \\
            RRN~\cite{li2018referring}     & DeepLab ResNet-101  & 55.33 & 57.26 & 53.95 & 39.75 & 42.15 & 36.11 & -     & -     \\
            CMSA~\cite{ye2019cross}    & DeepLab ResNet-101  & 58.32 & 60.61 & 55.09 & 43.76 & 47.6  & 37.89 & -     & -     \\
            CAC~\cite{chen2019referring}     & DeepLab ResNet-101  & 58.90 & 61.77 & 53.81 & -     & -     & -     & 46.37 & 46.95 \\
            STEP~\cite{chen2019see}    & DeepLab ResNet-101  & 60.04 & 63.46 & 57.97 & 48.19 & 52.33 & 40.41 & -     & -     \\
            BRINet~\cite{hu2020bi}  & DeepLab ResNet-101  & 60.98 & 62.99 & 59.21 & 48.17 & 52.32 & 42.11 & -     & -     \\
            CMPC~\cite{huang2020referring}    & DeepLab ResNet-101  & 61.36 & 64.53 & 59.64 & 49.56 & 53.44 & 43.23 & -     & -     \\
            LSCM~\cite{hui2020linguistic}    & DeepLab ResNet-101  & 61.47 & 64.99 & 59.55 & 49.34 & 53.12 & 43.50 & -     & -     \\
            CMPC+~\cite{liu2021cross}   & DeepLab ResNet-101  & 62.47 & 65.08 & 60.82 & 50.25 & 54.04 & 43.47 & -     & -     \\
            BUSNet~\cite{9577319}  & DeepLab ResNet-101  & 63.27 & 66.41 & 61.39 & 51.76 & 56.87 & 44.13 & -     & -     \\
            CGAN~\cite{Luo2020CascadeGA}    & DeepLab ResNet-101  & 64.86 & 68.04 & 62.07 & 51.03 & 55.51 & 44.06 & 51.01 & 51.69 \\
            EFN~\cite{feng2021encoder}     & Wide ResNet-101     & 62.76 & 65.69 & 59.67 & 51.50 & 55.24 & 43.01 & -     & -     \\
            ETRIS~\cite{xu2023bridging}   & CLIP ResNet-101     & 71.06 & 74.11 & 66.66 & 62.23 & 68.51 & 52.79 & 60.28 & 60.42 \\
            ReSTR~\cite{kim2022restr}   & ViT-B-16            & 67.22 & 69.3  & 64.45 & 55.78 & 60.44 & 48.27 & 54.48 & -     \\
            ETRIS~\cite{xu2023bridging}   & ViT-B-16            & 70.51 & 73.51 & 66.63 & 60.10 & 66.89 & 50.17 & 59.82 & 59.91 \\
            CRIS~\cite{wang2022cris}    & CLIP ResNet-50      & 69.52 & 72.72 & 64.7  & 61.39 & 67.1  & 52.48 & 59.87 & 60.36 \\
            CRIS~\cite{wang2022cris}    & CLIP ResNet-101     & 70.47 & 73.18 & 66.10 & 62.27 & 68.08 & 53.68 & 59.87 & 60.36 \\
            LAVT~\cite{yang2022lavt}    & Swin-B              & 72.73 & 75.82 & 68.79 & 62.14 & 68.38 & 55.10 & 61.24 & 62.10 \\
            \methodname (Ours)    & Mamba-B             & \bf74.54 & \bf76.74 & \bf70.89 & \bf65.00 & \bf70.78 & \bf57.53 &\bf63.9 &\bf 64.0                  \\
            \bottomrule
        \end{tabular}}
\end{table}

\cref{tab:main_table} shows the comparison of our \methodname with state-of-the-art methods on RefCOCO, RefCOCO+, and G-Ref datasets.
\methodname outperforms all other methods across all datasets, demonstrating its superior capability in capturing and integrating contextual information for more accurate segmentation.
In particular, Mamba outperforms previous Swin-based methods such as LAVT~\cite{yang2022lavt} by a considerable margin, 
indicating the remarkable capability of the Mamba-based architecture in segmentation tasks.
and the efficiency of the newly introduced Mamba Twister.%

\begin{table}[ht]
    \centering
    \caption{\textbf{Ablation on the combination of two scans.}
    We ablated each scan separately, as well as combining them parallel or swap the order.
    The occurence of Spatial Scan affect most to the performance. 
    The three variants performs similarly, with the combination of Spatial-Channel slightly better. Results are evaluated on RefCOCO.}
    \label{tab:ablation_scan}
    \setlength\tabcolsep{3pt}
    \resizebox{1.\linewidth}{!}{%
    \begin{tabular}{cc|cccc|cccc|cccc}
    \toprule
    & 
    &\multicolumn{4}{c|}{val}
    &\multicolumn{4}{c|}{testA}
    &\multicolumn{4}{c}{testB}\\
    Channel Scan & Spatial Scan 
    & mIoU & oIoU & Pr@50 & Pr@70 %
    & mIoU & oIoU & Pr@50 & Pr@70 %
    & mIoU & oIoU & Pr@50 & Pr@70 %
    \\
    \midrule
    \checkmark & - & 
    62.3 & 60.1 & 70.7 & 57.2 &
    64.9 & 63.2 & 74.3 & 61.9 &
    59.7 & 57.5 & 66.3 & 51.9 \\

    - & \checkmark & 
    70.0 & 65.8 & 80.5 & 69.1 &
    72.3 & 71.0 & 83.7 & 73.4 &
    67.5 & 63.5 & 75.2 & 62.7 \\
    
    \multicolumn{2}{c|}{Parallel} & 
    71.0 & 68.8 & 81.9 & 70.0 &
    73.1 & \bf 71.8 & 84.7 & 74.1 &
    67.9 & 64.6 & 76.9 & 64.5 \\
    
    \multicolumn{2}{c|}{Spatial-Channel} & 
    71.4 & \bf 68.8 & \bf 82.4 & 70.8 &
    73.3 & 71.4 & \bf 84.9 & 74.2 &
    67.9 & \bf 64.7 & 76.5 & 64.4 \\

    \multicolumn{2}{c|}{Channel-Spatial} & 
    \bf 71.6 &  68.4 & 82.1 & \bf 70.9 &
    \bf 73.3 & 71.5 & 84.8 & \bf 74.5 &
    \bf 68.4 & 64.5 & \bf 77.1 & \bf 64.9 \\
    \bottomrule
    \end{tabular}}
    \vspace{-1.5em}
\end{table}

\subsection{Ablation Study}

\subsubsection{Effects of Two Scans. }

\cref{tab:ablation_scan} presents a comparative analysis of different scan variants. 
We provide outcomes of conducting Channel Scan and Spatial Scan independently.
Additionally, results from various combinations of these two scans are discussed. 
The term ``Parallel'' denotes the concurrent execution of both scans, followed by adding the independent output for fusion. 
``Channel-Spatial'' refers to the original scanning order in Mamba Twister shown in~\cref{fig:architecture}, where the Channel Scan is executed first, followed by the Spatial Scan.
``Spatial-Channel'' refers to reversing the sequence of the two scans, starting with Spatial Scan and then proceeding to Channel Scan.

\cref{tab:ablation_scan} indicates that utilizing any single scan on its own yields suboptimal results, 
with the use of Channel Scan alone experiencing the most significant drop in effectiveness. 
This decline may indicate that a pure Channel Scan alters the data distribution, adversely affecting network stability. 
Among the three hybrid scanning policies, each has its strengths and weaknesses. 
Overall, the combination of Channel-Spatial Scan appears to offer a considerable advantage.

\subsubsection{Distribution Visualization. } 
\begin{figure}[ht]
    \centering
    \begin{subfigure}[b]{0.3\textwidth}
        \centering
        \includegraphics[width=0.7\textwidth]{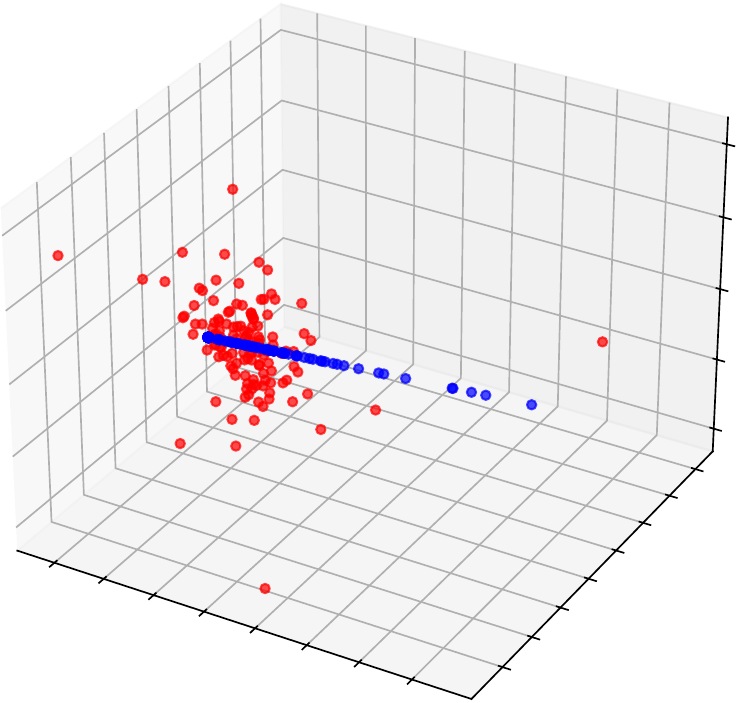}
        \caption{Input data distribution.}
        \label{fig:pca_vis1}
    \end{subfigure}
    \hfill
    \begin{subfigure}[b]{0.3\textwidth}
        \centering
        \includegraphics[width=0.7\textwidth]{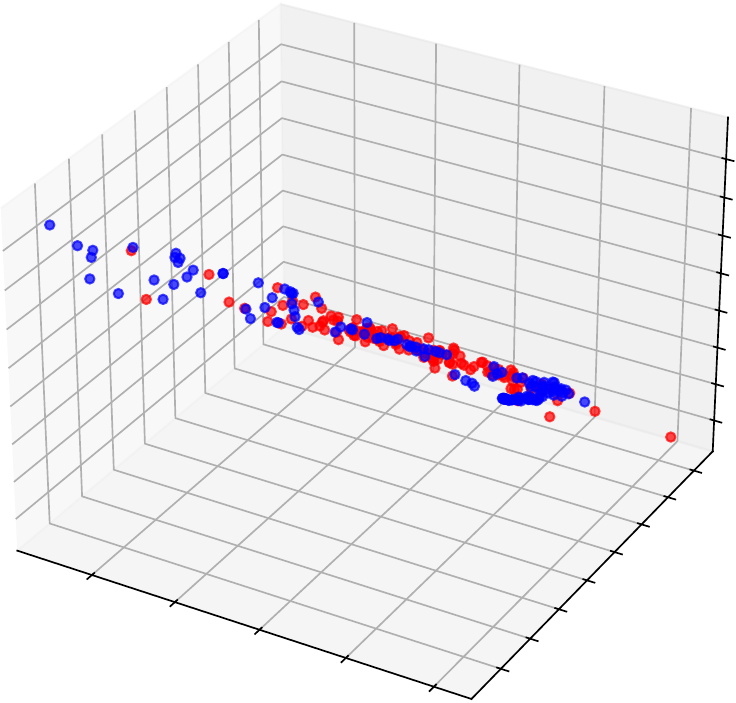}
        \caption{After Channel Scan.}
        \label{fig:pca_vis2}
    \end{subfigure}
    \hfill
    \begin{subfigure}[b]{0.3\textwidth}
        \centering
        \includegraphics[width=0.7\textwidth]{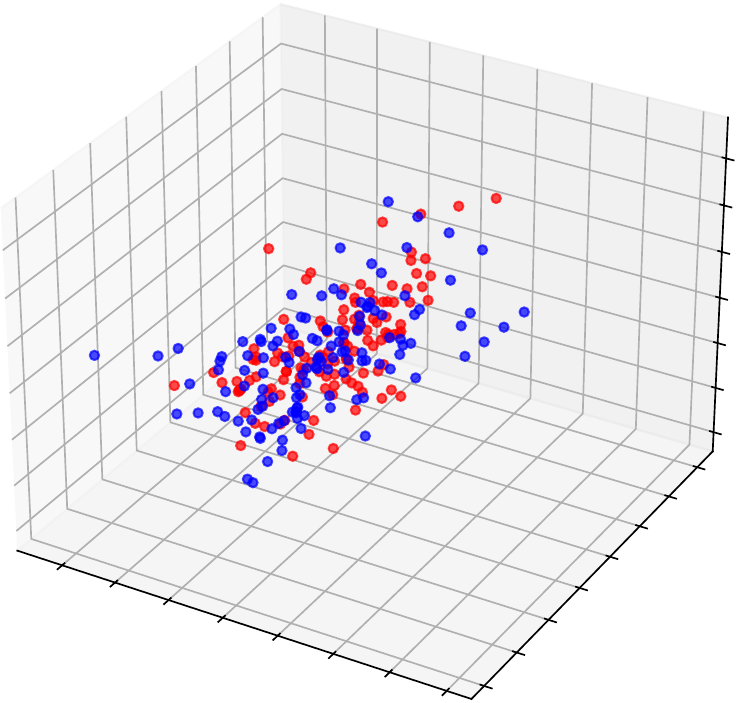}
        \caption{After Spatial Scan.}
        \label{fig:pca_vis3}
    \end{subfigure}
    \caption{\textbf{Data distribution after Channel Scan and Spacial Scan.} \textcolor{red}{\textbf{Image}} in red and \textcolor{blue}{\textbf{text}} data in blue. The Channel Scan tends to aggregate different modalities towards the distribution of textual side. The Spatial Scan reintegrates the previously aligned modalities, distributing them in a manner that reflects their combined influence.}
    \label{fig:pca_vis}
    \vspace{-1.5em}
\end{figure}

The visualization of the distribution of the multi-modal input data is depicted in~\cref{fig:pca_vis}, where we use PCA to map the feature to a 3D space. The \textcolor{red}{\textbf{image}} and \textcolor{blue}{\textbf{text}} data are indicated in red and blue respectively.

\cref{fig:pca_vis1} illustrates the distribution of the input data. We can observe that the text data forms a linear arrangement within the 3D space, as it is duplicated $H\times W$ times to align with the image size.
This suggests that the text instruction should be applied for each pixel within the image. 
\cref{fig:pca_vis2} reveals the data distribution following the application of a Channel Scan. 
Notably, this process appears to lean towards aggregating different modalities towards a distribution pattern similar with text. 
\cref{fig:pca_vis3}  demonstrates the data distribution after a Spatial Scan.
Here, the features of the textual and image data disperse, with the modalities intermixed. 
The Spatial Scan thus reintegrates the previously aligned modalities, distributing them in a manner that reflects their combined influence.

\subsubsection{Effects of Global and Local Features.}

\begin{table}[ht]
    \centering
    \caption{\textbf{Ablation study on global and local interaction.} Both global and local interactions are crucial for modality fusion. Results are evaluated on RefCOCO.
    }
    \label{tab:ablation_feature}
    \setlength\tabcolsep{3pt}
    \resizebox{1.\linewidth}{!}{%
    \begin{tabular}{ccc|cccc|cccc|cccc}
    \toprule
    & &
    &\multicolumn{4}{c|}{val}
    &\multicolumn{4}{c|}{testA}
    &\multicolumn{4}{c}{testB}\\
    Image & Global & Local 
    & mIoU & oIoU & Pr@50 & Pr@70 %
    & mIoU & oIoU & Pr@50 & Pr@70 %
    & mIoU & oIoU & Pr@50 & Pr@70 %
    \\
    \midrule
    \checkmark & \checkmark & - &
    69.1 & 66.6 & 79.5 & 66.3 &
    71.4 & 69.8 & 82.3 & 70.9 &
    66.3 & 63.7 & 74.7 & 61.0 \\

    \checkmark & - & \checkmark &
    69.9 & 67.9 & 80.6 & 68.2 &
    72.2 & 70.6 & 83.9 & 73.1 &
    66.4 & 63.8 & 75.3 & 61.7 \\
    
    \checkmark & \checkmark&\checkmark & 
    \bf 71.6 & \bf 68.4 & \bf 82.1 & \bf 70.9 &
    \bf 73.3 & \bf 71.5 & \bf 84.8 & \bf 74.5 &
    \bf 68.4 & \bf 64.5 & \bf 77.1 & \bf 64.9 \\

    \bottomrule
    \end{tabular}}
    \label{tab:feature_ablation}
    \vspace{-1.5em}
\end{table}

We also ablate the effect of global and local interactions when forming the hybrid feature cube, formally, the $\tilde{\mathbf{F}}_t$ and $\tilde{\mathbf{F}}_c$ in~\cref{eq:hybrid_cube}.
The results in~\cref{tab:ablation_feature} illustrate that the integration of both global and local features significantly enhances the performance. 

\section{Visualization Results}
\begin{figure}
    \vspace{-3em}
    \includegraphics[width=0.16\textwidth]{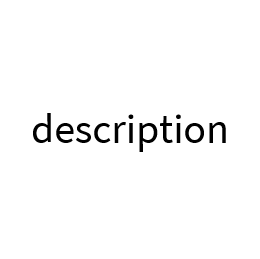}
    \includegraphics[width=0.16\textwidth]{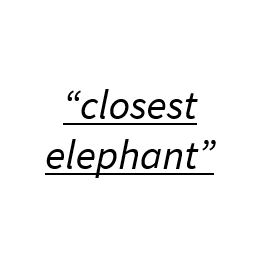}
    \includegraphics[width=0.16\textwidth]{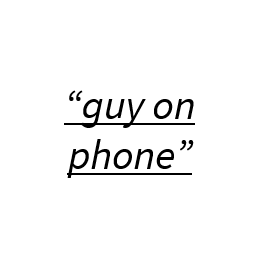}
    \includegraphics[width=0.16\textwidth]{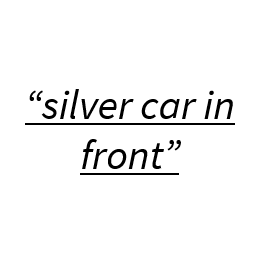}
    \includegraphics[width=0.16\textwidth]{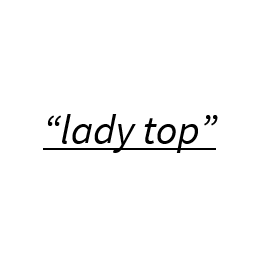}
    \includegraphics[width=0.16\textwidth]{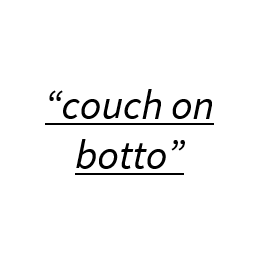}
    \includegraphics[width=0.16\textwidth]{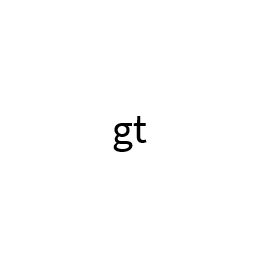}
    \includegraphics[width=0.16\textwidth]{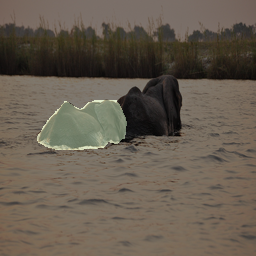}
    \includegraphics[width=0.16\textwidth]{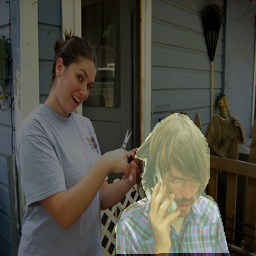}
    \includegraphics[width=0.16\textwidth]{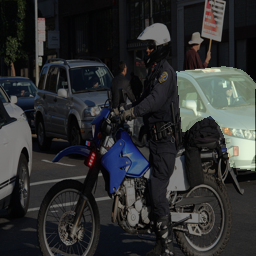}
    \includegraphics[width=0.16\textwidth]{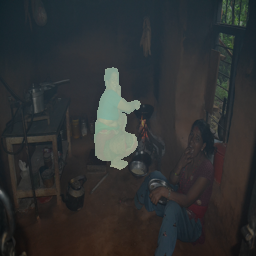}
    \includegraphics[width=0.16\textwidth]{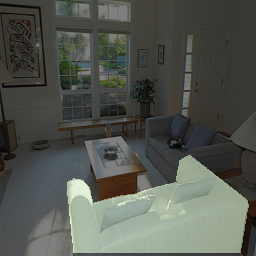}
    \includegraphics[width=0.16\textwidth]{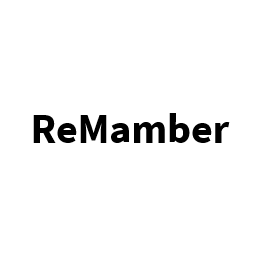}
    \includegraphics[width=0.16\textwidth]{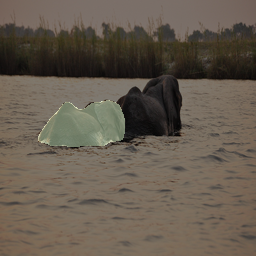}
    \includegraphics[width=0.16\textwidth]{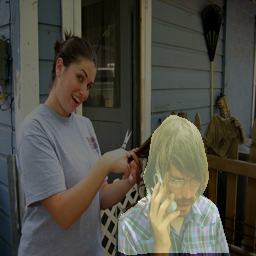}
    \includegraphics[width=0.16\textwidth]{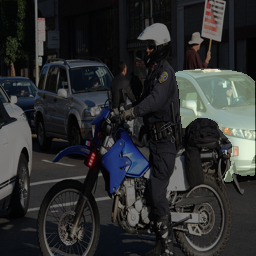}
    \includegraphics[width=0.16\textwidth]{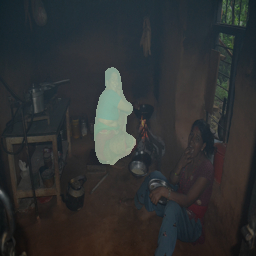}
    \includegraphics[width=0.16\textwidth]{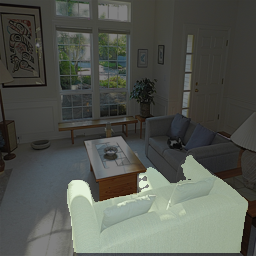}
    \includegraphics[width=0.16\textwidth]{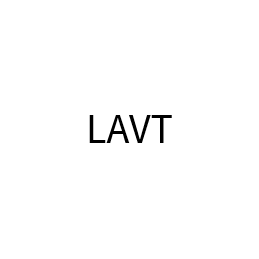}
    \includegraphics[width=0.16\textwidth]{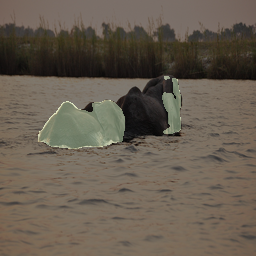}
    \includegraphics[width=0.16\textwidth]{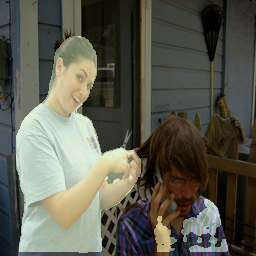}
    \includegraphics[width=0.16\textwidth]{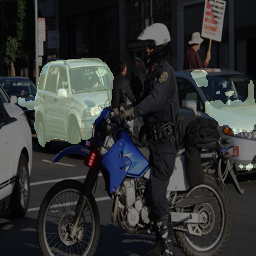}
    \includegraphics[width=0.16\textwidth]{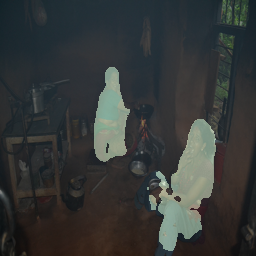}
    \includegraphics[width=0.16\textwidth]{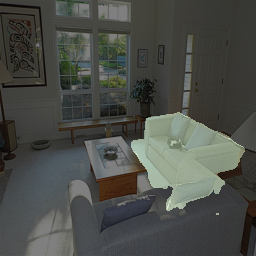}
    \caption{Visualization results of our \textbf{ReMamber} and the baseline model LAVT. Our model is able to predict more accurate masks.}
    \label{fig:vis1}
\end{figure}
\begin{figure}
    \includegraphics[width=0.16\textwidth]{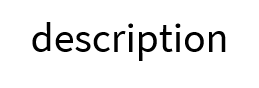}
    \includegraphics[width=0.16\textwidth]{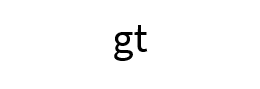}
    \includegraphics[width=0.16\textwidth]{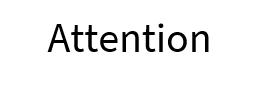}
    \includegraphics[width=0.16\textwidth]{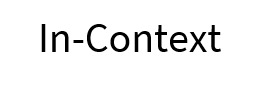}
    \includegraphics[width=0.16\textwidth]{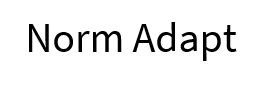}
    \includegraphics[width=0.16\textwidth]{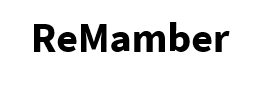}
    \includegraphics[width=0.16\textwidth]{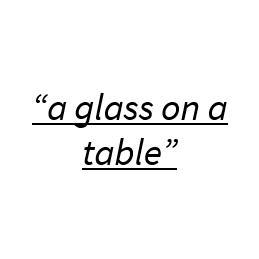}
    \includegraphics[width=0.16\textwidth]{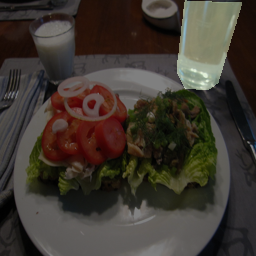}
    \includegraphics[width=0.16\textwidth]{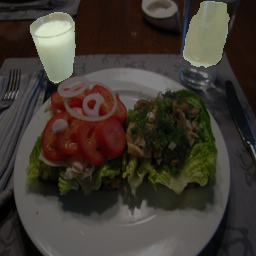}
    \includegraphics[width=0.16\textwidth]{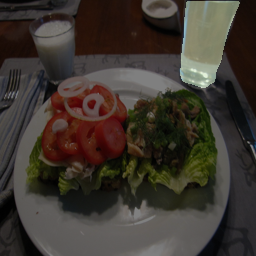}
    \includegraphics[width=0.16\textwidth]{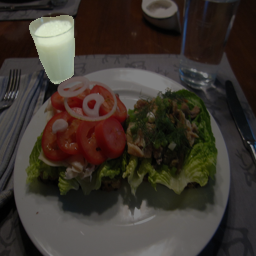}
    \includegraphics[width=0.16\textwidth]{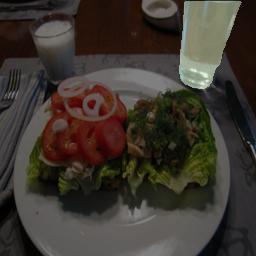}
    \includegraphics[width=0.16\textwidth]{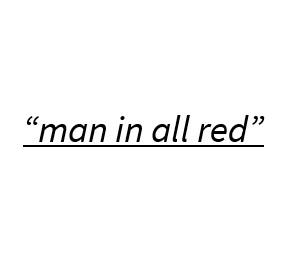}
    \includegraphics[width=0.16\textwidth]{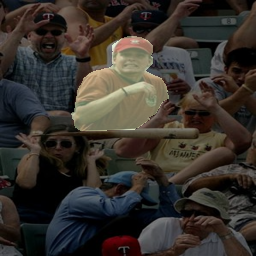}
    \includegraphics[width=0.16\textwidth]{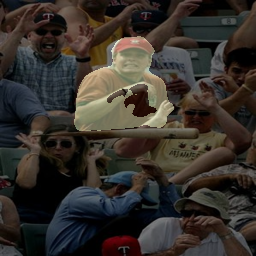}
    \includegraphics[width=0.16\textwidth]{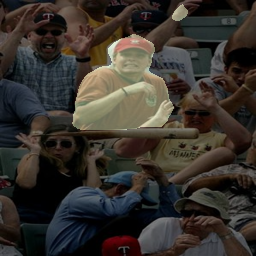}
    \includegraphics[width=0.16\textwidth]{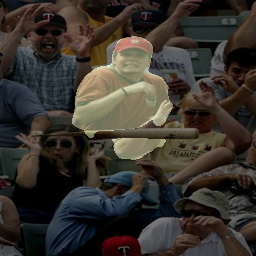}
    \includegraphics[width=0.16\textwidth]{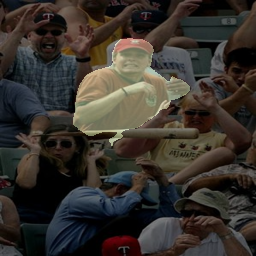}
    \includegraphics[width=0.16\textwidth]{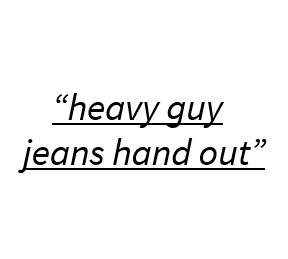}
    \includegraphics[width=0.16\textwidth]{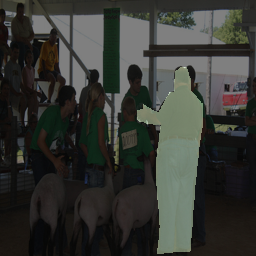}
    \includegraphics[width=0.16\textwidth]{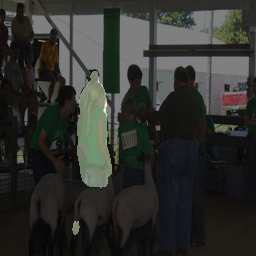}
    \includegraphics[width=0.16\textwidth]{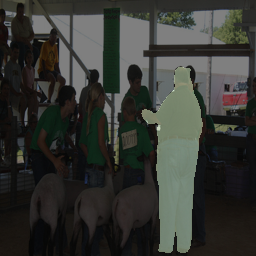}
    \includegraphics[width=0.16\textwidth]{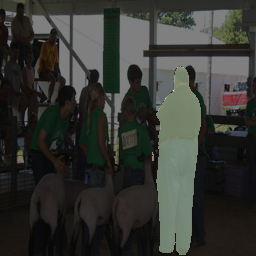}
    \includegraphics[width=0.16\textwidth]{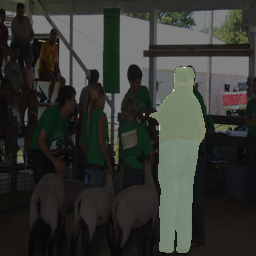}
    \includegraphics[width=0.16\textwidth]{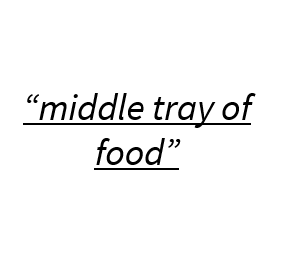}
    \includegraphics[width=0.16\textwidth]{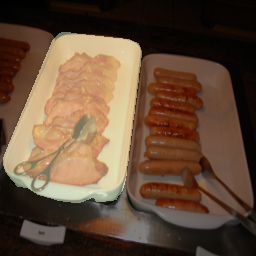}
    \includegraphics[width=0.16\textwidth]{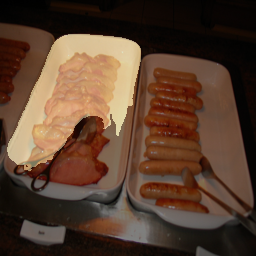}
    \includegraphics[width=0.16\textwidth]{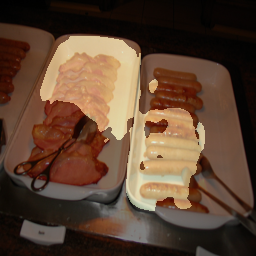}
    \includegraphics[width=0.16\textwidth]{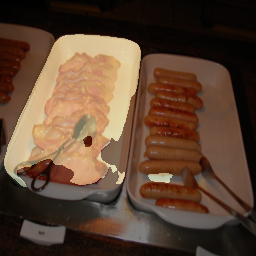}
    \includegraphics[width=0.16\textwidth]{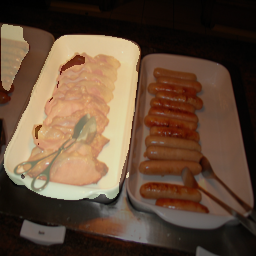}
    \includegraphics[width=0.16\textwidth]{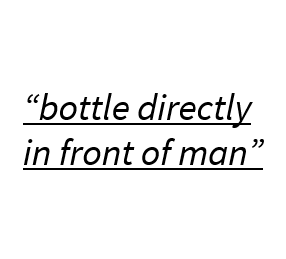}
    \includegraphics[width=0.16\textwidth]{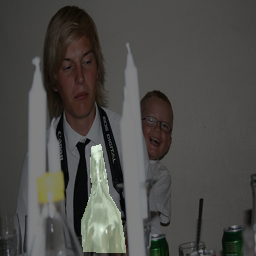}
    \includegraphics[width=0.16\textwidth]{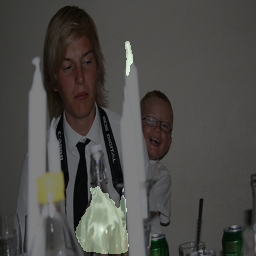}
    \includegraphics[width=0.16\textwidth]{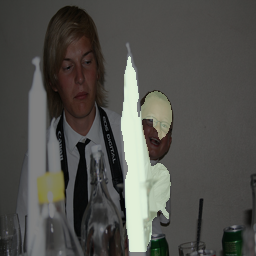}
    \includegraphics[width=0.16\textwidth]{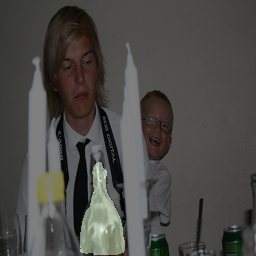}
    \includegraphics[width=0.16\textwidth]{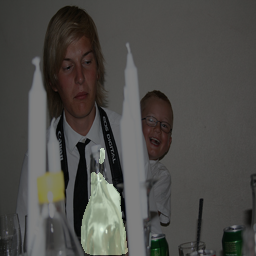}
    \includegraphics[width=0.16\textwidth]{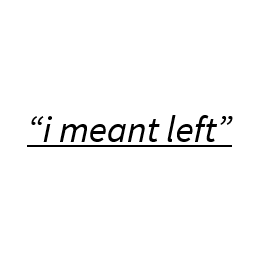}
    \includegraphics[width=0.16\textwidth]{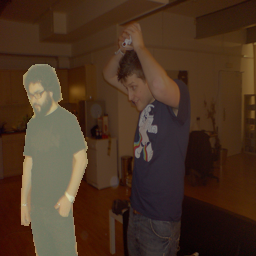}
    \includegraphics[width=0.16\textwidth]{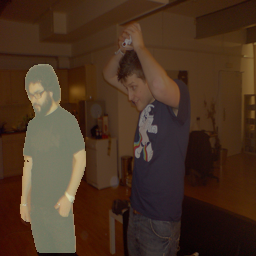}
    \includegraphics[width=0.16\textwidth]{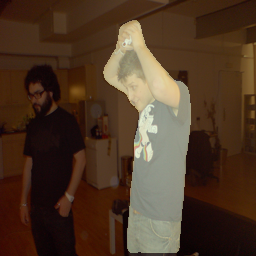}
    \includegraphics[width=0.16\textwidth]{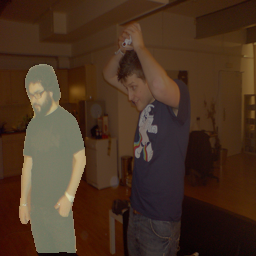}
    \includegraphics[width=0.16\textwidth]{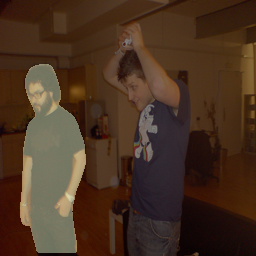}
    \includegraphics[width=0.16\textwidth]{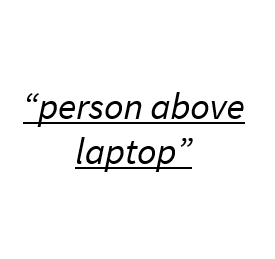}
    \includegraphics[width=0.16\textwidth]{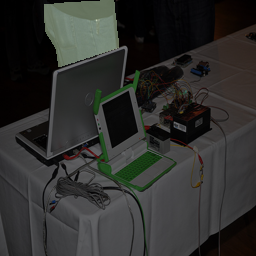}
    \includegraphics[width=0.16\textwidth]{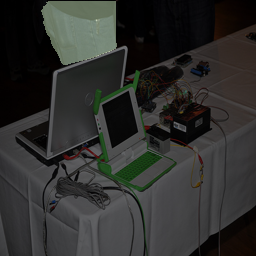}
    \includegraphics[width=0.16\textwidth]{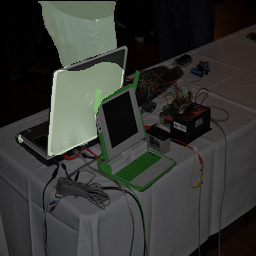}
    \includegraphics[width=0.16\textwidth]{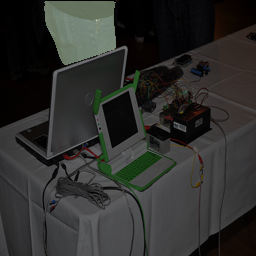}
    \includegraphics[width=0.16\textwidth]{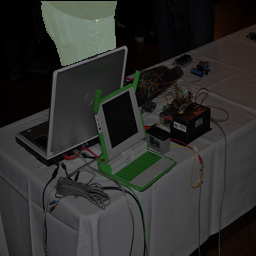}
    \includegraphics[width=0.16\textwidth]{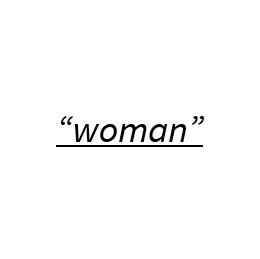}
    \includegraphics[width=0.16\textwidth]{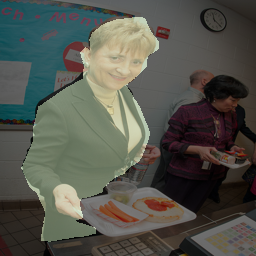}
    \includegraphics[width=0.16\textwidth]{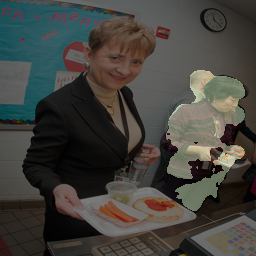}
    \includegraphics[width=0.16\textwidth]{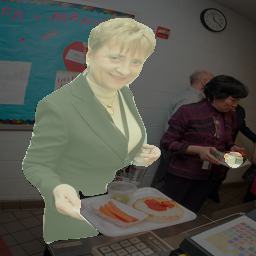}
    \includegraphics[width=0.16\textwidth]{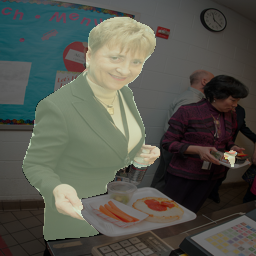}
    \includegraphics[width=0.16\textwidth]{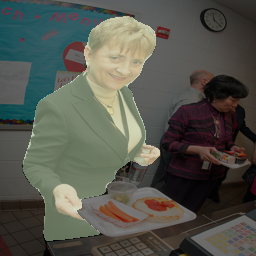}
    \includegraphics[width=0.16\textwidth]{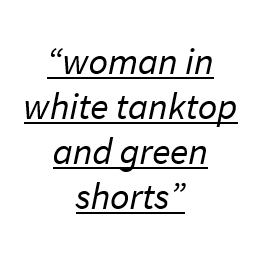}
    \includegraphics[width=0.16\textwidth]{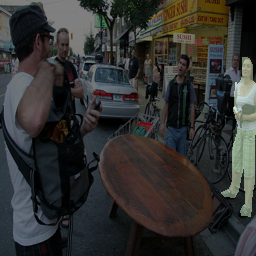}
    \includegraphics[width=0.16\textwidth]{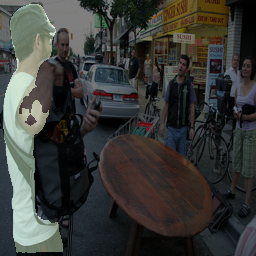}
    \includegraphics[width=0.16\textwidth]{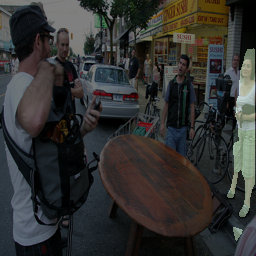}
    \includegraphics[width=0.16\textwidth]{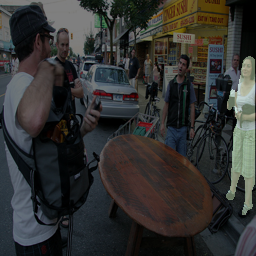}
    \includegraphics[width=0.16\textwidth]{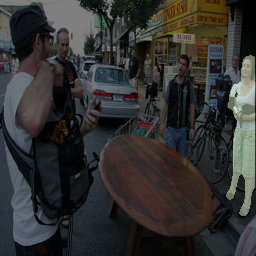}
    \caption{\textbf{Visualization results.} Our \textbf{ReMamber} is capable of producing segmentation results with higher accuracy. 
While other three variants occasionally encounter issues with inaccurate segmentation masks or are misled towards incorrect objects.}
    \label{fig:vis}
\end{figure}

\cref{fig:vis1} presents the visualization results of our method compared with the baseline method LAVT~\cite{yang2022lavt}. \cref{fig:vis} presents the visualization outcomes of our method alongside three other variants. 
Our \textbf{ReMamber} is capable of producing segmentation results with higher accuracy. 
In contrast, the other three variants occasionally encounter issues with inaccurate segmentation masks or are misled towards incorrect objects.

\section{Conclusion}

In this study, we introduce \methodname, a novel architecture utilizing the Mamba framework in Referring Image Segmentation (RIS).
It marks a significant advancement in multi-modal understanding. By integrating visual and textual information through innovative Mamba Twister blocks, our approach sets new benchmarks in image-text modality fusion. Achieving competitive results across multiple RIS datasets, our research highlights the potential of Mamba architecture in enhancing the scalability and performance of multi-modal tasks, offering promising directions for future exploration in the field.

\subsubsection{Limitations and Future Works.}

In our current architecture, the segmentation decoder is constructed by only a few convolutional layers. 
As shown in~\cref{sec:CI_ablation}, the integration of the Cross-Attention mechanism within the Mamba-based architecture demonstrates sub-optimal compatibility, undermining the overall efficacy of our method. 
In light of these findings, future endeavors will be directed towards the investigation of more sophisticated multi-modal segmentation decoders which best fit Mamba architecture.

\appendix
\section*{Appendix}

\section{Speed Analysis}
\begin{figure}
    \centering
    \includegraphics[width=0.85\linewidth]{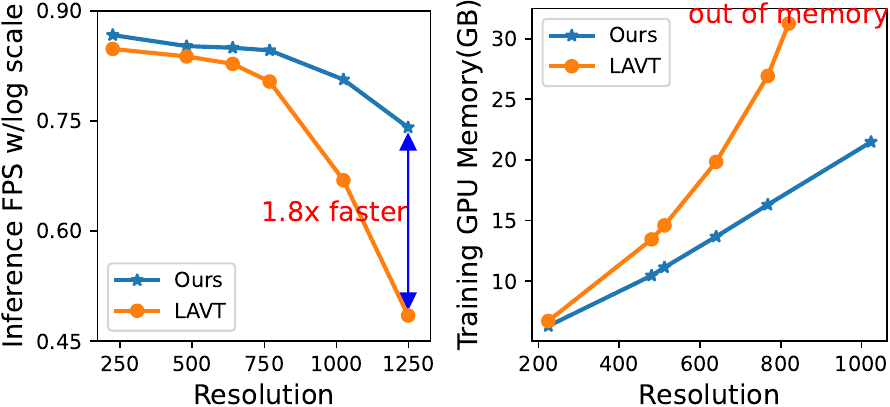}
    \caption{Comparison of inference FPS (left) and training GPU memory (right) between LAVT and our \methodname.}
    \vspace{-3em}
    \label{fig:efficiency}
\end{figure}

Here, we supplement two experiments \textit{w.r.t.} model FPS and memory cost in Fig.~\ref{fig:efficiency} under different resolutions. ReMamber is consistently faster and requires fewer memory cost than LAVT, especially with large resolution (\textit{e.g.}, 1,024). 

\section{Implementation Details}
Here we provide more details about the implementation of our method, including detailed architectural structure, training settings and other baseline implementation.

\subsection{Architecture Details}
The whole \methodname architecture consists of an encoder and a decoder. 
The encoder is made up by a patch-embedding layer with patch-size 4 and hidden dimension 128, followed by 4 Mamba Twister blocks.
Each Twister block consists of several VSS Layers and a Twisting Layer. 
The VSS Layer number configuration is set as 2-2-15-2, with hidden dimension 128-256-512-1024, respectively.

For the decoder part, we provide two variants in our code implementation: convolution-based decoder (\texttt{ReMamber\_Conv}) and Mamba-based decoder (\texttt{ReMamber\_Mamba}).
\texttt{ReMamber\_Conv} uses a progressive upsampling architecture with 4 residual blocks, 2 convolutional layers in each.
\texttt{ReMamber\_Mamba} is similar with \texttt{ReMamber\_Conv}, but uses VSS layers instead of convolutional layers. This variant is slightly faster.

\subsection{Implementation for Other Three Variants}
Here, we detail the implementation of other three architecture variants in our paper, \ie, In-context Conditioning, Attention-based Conditioning and Norm Adaptation.

\noindent\textbf{In-context Conditioning. }
To enable the model to distinguish between two modalities, we add learnable positional embeddings to the image and text tokens separately at each layer before the Spatial SSM.

\noindent\textbf{Attention-based Conditioning. }
In this variant, we also incorporate learnable positional embeddings. For the cross-attention block, we use image tokens as the query, and text tokens as the key and value.

\noindent\textbf{Norm Adaptation. }
Norm Adaptation learns a global scale and bias. First, we use an MLP layer to pool a global vector from the text. This vector is then used to scale and bias the image tokens. An additional feed-forward layer is added after adjusting the scale and bias to maintain parameter size comparable to other variants.

\clearpage  %

\bibliographystyle{splncs04}
\bibliography{main}
\end{document}